%% file: hashnn_main.tex
\documentclass{article}

\usepackage{times}
\usepackage{graphicx} % more modern
\usepackage{subfigure} 

\usepackage{natbib}

\usepackage{algorithm}
\usepackage{algorithmic}

\usepackage{hyperref}

\usepackage{icml2015.arxiv}

\usepackage{amsmath}
\usepackage{amssymb}
\usepackage{amsfonts}
\usepackage{color}

\newcommand{\fullname}[0]{Hashed Neural Nets}
\newcommand{\abbrev}[0]{HashedNets}
\newcommand{\abbrevtable}[0]{HashNet}
\newcommand{\abbrevtabledk}[0]{HashNet$_{\textrm{DK}}$}
\newcommand{\ba}[0]{{\textbf a}}

\newcommand{\bx}[0]{{\textbf x}}

\newcommand{\bV}[0]{{\textbf V}}

\newcommand{\bw}[0]{{\textbf w}}

\newcommand{\bz}[0]{{\textbf z}}
\newcommand{\mR}[0]{\mathcal{R}}
\newcommand{\boldBlue}[1]{\textcolor{blue}{\textbf{#1}}}

\newcommand{\grad}[2]{ \frac{\partial {#1}}{\partial #2} }
\newcommand{\gradL}[1]{ \frac{\partial \cal{L}}{\partial #1} }

\newcommand{\hij}[0]{h^\ell(i,j)}
\newcommand{\hijnol}{h(i,j)}
\newcommand{\hh}[2]{h^\ell(#1,#2)}
\newcommand{\hhl}[3]{h^{#1}(#2,#3)}
\newcommand{\xij}[0]{\xi^\ell(i,j)}
\newcommand{\xx}[2]{\xi^\ell(#1,#2)}

\icmltitlerunning{Compressing Neural Networks with the Hashing Trick}

\begin{document} 

\twocolumn[
\icmltitle{Compressing Neural Networks with the Hashing Trick}
%HashedNets: Compressed Neural Networks with Finite Memory}

% unless you've provided the [accepted] option to the icml2015
\icmlauthor{Wenlin Chen$^*$}{wenlinchen@wustl.edu}
\icmlauthor{James T. Wilson$^*$}{j.wilson@wustl.edu}
\icmlauthor{Stephen Tyree$^{*\dag}$}{styree@nvidia.com}
\icmlauthor{Kilian Q. Weinberger$^*$}{kilian@wustl.edu}
\icmlauthor{Yixin Chen$^*$}{chen@cse.wustl.edu}
\icmladdress{
	$^*$
	Department of Computer Science and Engineering,
	Washington University in St. Louis,
	St. Louis, MO, USA \\
	$^\dag$
	NVIDIA, Santa Clara, CA, USA
}

\icmlkeywords{feature hashing, neural network, deep learning, compression}

\vskip 0.3in
]

\begin{abstract} 
\input content/abstract.tex
\end{abstract} 

\input{content/intro.tex}
\input{content/background.tex}

\input{content/notation.tex}
\input{content/method.tex}

\input{content/featurehashing.tex}

\input{content/backprop.tex}

\input{content/related.tex}
\input{content/experiment.tex}

\input{content/discussion.tex}

\bibliography{hashnn,kilian}
\bibliographystyle{icml2015}

\end{document}

%% file: content/abstract.tex
%!TEX root=../hashnn_main.tex
%
%Recent breakthroughs in deep learning have improved voice recognition and object classification to accuracy levels sufficient for everyday consumer products, such as smart phones or medical devices. 
As deep nets are increasingly used in applications suited for mobile devices, a fundamental dilemma becomes apparent: the trend in deep learning is to grow models to absorb ever-increasing data set sizes; however mobile devices are designed with very little memory and cannot store such large models. 
We present a novel network architecture, \abbrev{}, that exploits inherent redundancy in neural networks to achieve drastic reductions in model sizes. \abbrev{} uses a low-cost hash function to randomly group connection weights into hash buckets, and all connections within the same hash bucket share a single parameter value. These parameters are tuned to adjust to the \abbrev{} weight sharing architecture with standard backprop during training. Our hashing procedure introduces no additional memory overhead, and we demonstrate on several benchmark data sets that \abbrev{} shrink the storage requirements of neural networks substantially while mostly preserving generalization performance. 

%% file: content/intro.tex
%!TEX root=../hashnn_main.tex

\section{Introduction}

% Deep learning has grown in size resulting in huge networks
In the past decade deep neural networks have set new performance standards in many high-impact applications.
These include object classification~\cite{krizhevsky2012imagenet,sermanet2013overfeat}, speech recognition~\cite{hinton2012deep}, image caption generation~\cite{vinyals2014show,karpathy2014deep} and domain adaptation~\cite{glorot2011domain}.
As data sets increase in size, so do the number of parameters in these neural networks in order to absorb the enormous amount of supervision~\cite{coates2013deep}.  Increasingly, these networks are trained on industrial-sized clusters~\cite{le2013building} or high-performance graphics processing units (GPUs)~\cite{coates2013deep}.

% at the same time, computing platforms shrink in size
Simultaneously, there has been a second trend as applications of machine learning have shifted toward mobile and embedded devices.
As examples, modern smart phones are increasingly operated through speech recognition~\cite{schuster2010speech}, robots and self-driving cars perform object recognition in real time~\cite{montemerlo2008junior}, and medical devices collect and analyze patient data~\cite{lee2013low}.
In contrast to GPUs or computing clusters, these devices are designed for low power consumption and long battery life.
Most importantly, they typically have small working memory.
For example, even the top-of-the-line iPhone 6 only features a mere 1GB of  RAM.\footnote{\url{http://en.wikipedia.org/wiki/IPhone_6}}

% this creates a dilemma
The disjunction between these two trends creates a dilemma when state-of-the-art deep learning algorithms are designed for deployment on mobile devices.
While it is possible to train deep nets offline on industrial-sized clusters (server-side), the sheer size of the most effective models would exceed the available memory, making it prohibitive to perform testing on-device.
In speech recognition, one common cure is to transmit processed voice recordings to a computation center, where the voice recognition is performed server-side~\cite{chun2009}. 
This approach is problematic, as it only works when sufficient bandwidth is available and incurs artificial delays through network traffic \cite{forbes2012}.
One solution is to train small models for the on-device classification; however, these tend to significantly impact accuracy \cite{chun2009}, leading to customer frustration.  

% Dark Knowledge
This dilemma motivates \emph{neural network compression}.
Recent work by \citet{denil2013predicting} demonstrates that there is a surprisingly large amount of redundancy among the weights of neural networks.
The authors show that a small subset of the weights are sufficient to reconstruct the entire network.
They exploit this by training low-rank decompositions of the weight matrices. \citet{Caruana2014} show that deep neural networks can be successfully compressed into ``shallow'' single-layer neural networks by training the small network on the (log-) outputs of the fully trained deep network~\cite{bucilua2006model}. \citet{courbariaux2015low} train neural networks with reduced bit precision, and,  long predating this work, \citet{lecun1989optimal} investigated dropping unimportant weights in neural networks.
In summary, the accumulated evidence suggests that much of the information stored within network weights may be redundant.

% Our paper overview
In this paper we propose \emph{\abbrev{}}, a novel network architecture to reduce and limit the memory overhead of neural networks.
%Our approach is compellingly simple, yet effective.
%We use a hash function to group network connections into hash buckets uniformly at random.
%All connections within the $i^{th}$ hash bucket share the same weight value $p_i$.
Our approach is compellingly simple: we use a hash function to group network connections into hash buckets uniformly at random such that all connections grouped to the $i^{th}$ hash bucket share the same weight value $w_i$.
Our parameter hashing is akin to prior work in feature hashing~\cite{weinberger09feature,hashKernelShi:2009,ganchev:08} and is similarly fast and requires no additional memory overhead.
The backpropagation algorithm~\cite{lecun2012efficient} can naturally tune the hash bucket parameters and take into account the random weight sharing within the neural network architecture.

% Results
We demonstrate on several real world deep learning benchmark data sets that \abbrev{} can drastically reduce the model size of neural networks with little impact in prediction accuracy. 
Under the same memory constraint, \abbrev{} have more adjustable free parameters than the low-rank decomposition methods suggested by \citet{denil2013predicting}, leading to smaller drops in descriptive power. 

Similarly, we also show that for a finite set of parameters it is beneficial to ``inflate'' the network architecture by re-using each parameter value multiple times. Best results are achieved when networks are inflated by a factor $8$--$16\times$. %$8\!-\!16\times$.
The ``inflation'' of neural networks with \abbrev{} imposes no restrictions on other network architecture design choices, such as dropout regularization~\cite{srivastava2014dropout}, activation functions~\cite{glorot2011deep,lecun2012efficient}, or weight sparsity~\cite{coates2011analysis}. 

%% file: content/background.tex
\section{Feature Hashing}

Learning under memory constraints has previously been explored in the context of large-scale learning for sparse data sets. \emph{Feature hashing} (or the \emph{hashing trick})~\cite{weinberger09feature,hashKernelShi:2009} is a technique to map high-dimensional text documents directly into bag-of-word~\cite{salton1988term} vectors, which would otherwise require use of memory consuming dictionaries for storage of indices corresponding with specific input terms.

Formally, an input vector $\bx\in \mathcal{R}^d$ is mapped into a feature space with a mapping function $\phi\colon\!\mathcal{R}^d\rightarrow \mathcal{R}^k$ where $k\!\ll\! d$. The mapping $\phi$ is based on two (approximately uniform) hash functions  $h\colon\!\mathbb{N}\!\rightarrow\!\{1,\dots,k\}$ and $\xi\colon\!\mathbb{N}\!\rightarrow\!\{-1,+1\}$ and the $k^{th}$ dimension of the hashed input $\bx$ is defined as $\phi_k(\bx)=\sum_{i:h(i)=k}x_i\xi(i)$.

The hashing trick leads to large memory savings for two reasons: it can operate directly on the input term strings and avoids the use of a dictionary to translate words into vectors; and the parameter vector of a learning model lives within the much smaller dimensional $\mathcal{R}^k$ instead of $\mathcal{R}^d$. 
The dimensionality reduction comes at the cost of collisions, where multiple words are mapped into the same dimension. This problem is less severe for sparse data sets and can be counteracted through multiple hashing~\cite{hashKernelShi:2009} or larger hash tables~\cite{weinberger09feature}.

In addition to memory savings, the hashing trick has the appealing property of being sparsity preserving, fast to compute and storage-free. 
The most important property of the hashing trick is, arguably, its (approximate) preservation of inner product operations. The second hash function, $\xi$,  guarantees that inner products are unbiased in expectation~\cite{weinberger09feature}; that is,
\begin{equation}
	\mathbb{E}[\phi(\bx)^\top\phi(\bx^\prime)]_{\phi} = \bx^\top\bx^\prime. 
	\label{eq.inner_product_hash}
\end{equation}
Finally, \citet{weinberger09feature} also show that the hashing trick can be used to learn multiple classifiers within the same hashed space. In particular, the authors use it for multi-task learning and define multiple hash functions $\phi_1,\dots,\phi_T$, one for each task, that map inputs for their respective tasks into one joint space. 
Let $\bw_1,\dots,\bw_T$ denote the weight vectors of the respective learning tasks, then if $t'\neq t$ a classifier for task $t'$ does not interfere with a hashed input for task $t$; \emph{i.e.} $\bw_t^\top\phi_{t'}(\bx)\approx 0$.

%% file: content/notation.tex
%!TEX root=../hashnn_main.tex

\section{Notation}
Throughout this paper we type vectors in bold ($\bx$), scalars in regular ($C$ or $b$) and matrices in capital bold ($\mathbf{X}$). Specific entries in vectors or matrices are scalars and follow the corresponding convention, $i.e.$ the $i^{th}$ dimension of vector $\mathbf{x}$ is $x_i$ and the $(i,j)^{th}$ entry of matrix $\mathbf{V}$ is $V_{ij}$. 

\paragraph{Feed Forward Neural Networks.}
We define the forward propagation of the $\ell^{th}$ layer in a neural networks as, 
\begin{equation}
	a^{\ell+1}_i = f(z^{\ell+1}_i),  ~~~\text{where}~~ z^{\ell+1}_i = \sum_{j=0}^{n^\ell} V_{ij}^\ell a^\ell_j,\label{eq.output_std}
\end{equation}
where $\bV^\ell$ is the (virtual) weight matrix in the $\ell^{th}$ layer. The vectors $\bz^\ell, \ba^\ell\!\in\!{\cal R}^{n^\ell}$  denote the activation units before and after transformation through the transition function $f(\cdot)$. 
Typical activation functions are rectifier linear unit (ReLU)~\cite{nair2010rectified}, sigmoid or tanh~\cite{lecun2012efficient}. 

%Denote $\bx\!=\![x_1,\dots,x_D]^\top$ as an input to the neural network, where $D$ is the number of features. Let $M$ denote the number of layers in the neural network and $\ai{l}$ denote the activation of unit $i$ in layer $l$. For the input layer, we have $l\!=\!1$, $\ai{1}=x_i$ for $i=1,\dots,D$. Let $n^l$ be the number of hidden units in layer $l$ and $\al{l}=[a^l_0,\dots,a^l_{n^l}]^\top$ be the vector of activations of layer $l$, where $a^l_0=1$ is a dummy variable for bias. In a standard fully-connected neural network, the value of neurons in layer $l+1$ is an activation function of a weighted sum over neurons in layer $l$, as follows:
%Here, $\bW^l\in \mathcal{R}^{n^{l+1}\times (n^l+1)}$ is the weight matrix of layer $l$ and $W_{ij}^l$ is the element in $i^{th}$ row and $j^{th}$ column. $f(\cdot)$ is an activation function, such as Rectifier Linear Unit (ReLu), sigmoid, or tanh. The weight matrix $\bW$ has $n^{l+1}(n^l+1)$ parameters. These parameters represent the only model storage requirement and the major source of memory consumption during evaluation. Further, in deep networks, this storage cost can be significant.

%% file: content/method.tex
%!TEX root=../hashnn_main.tex
\section{HashedNets}

In this section we present \abbrev{}, a novel variation of neural networks with drastically reduced model sizes (and memory demands). 
We first introduce our approach as a method of random weight sharing across the network connections and then describe how to facilitate it with the hashing trick to avoid any additional memory overhead.

% \input{content/notation.tex}

% \subsection{Random Weight Sharing For Reducing Free Parameters}
\subsection{Random weight sharing}
In a standard fully-connected neural network, there are $(n^\ell+1)\times n^{\ell+1}$ weighted connections between a pair of layers, each with a corresponding free parameter in the weight matrix $\bV^\ell$. 
We assume a finite memory budget per layer, $K^\ell\ll (n^\ell+1)\times n^{\ell+1}$, that cannot be exceeded. The obvious solution is to fit the neural network within budget by reducing the number of nodes $n^\ell, n^{\ell+1}$ in layers $\ell,\ell+1$ or by reducing the bit precision of the weight matrices~\cite{courbariaux2015low}. However if $K^\ell$ is sufficiently small, both approaches significantly reduce the ability of the neural network to generalize (see Section~\ref{sec:exp}). Instead, we propose an alternative: we keep the size of $\bV^\ell$ untouched but reduce its \emph{effective} memory footprint through \emph{weight sharing}. We only allow exactly $K^\ell$ different weights to occur within $\bV^\ell$, which we store in a weight vector $\bw^\ell \!\in\! \mathcal{R}^{K^\ell}$. The weights within $\bw^\ell$ are shared across multiple randomly chosen connections within $\bV^\ell$. 
We refer to the resulting matrix $\bV^\ell$ as \emph{virtual}, as its size could be increased (\emph{i.e.} nodes are added to hidden layer) without increasing the \emph{actual} number of parameters of the neural network.

\begin{figure}
    \centerline{
        \includegraphics[width=0.5\textwidth]{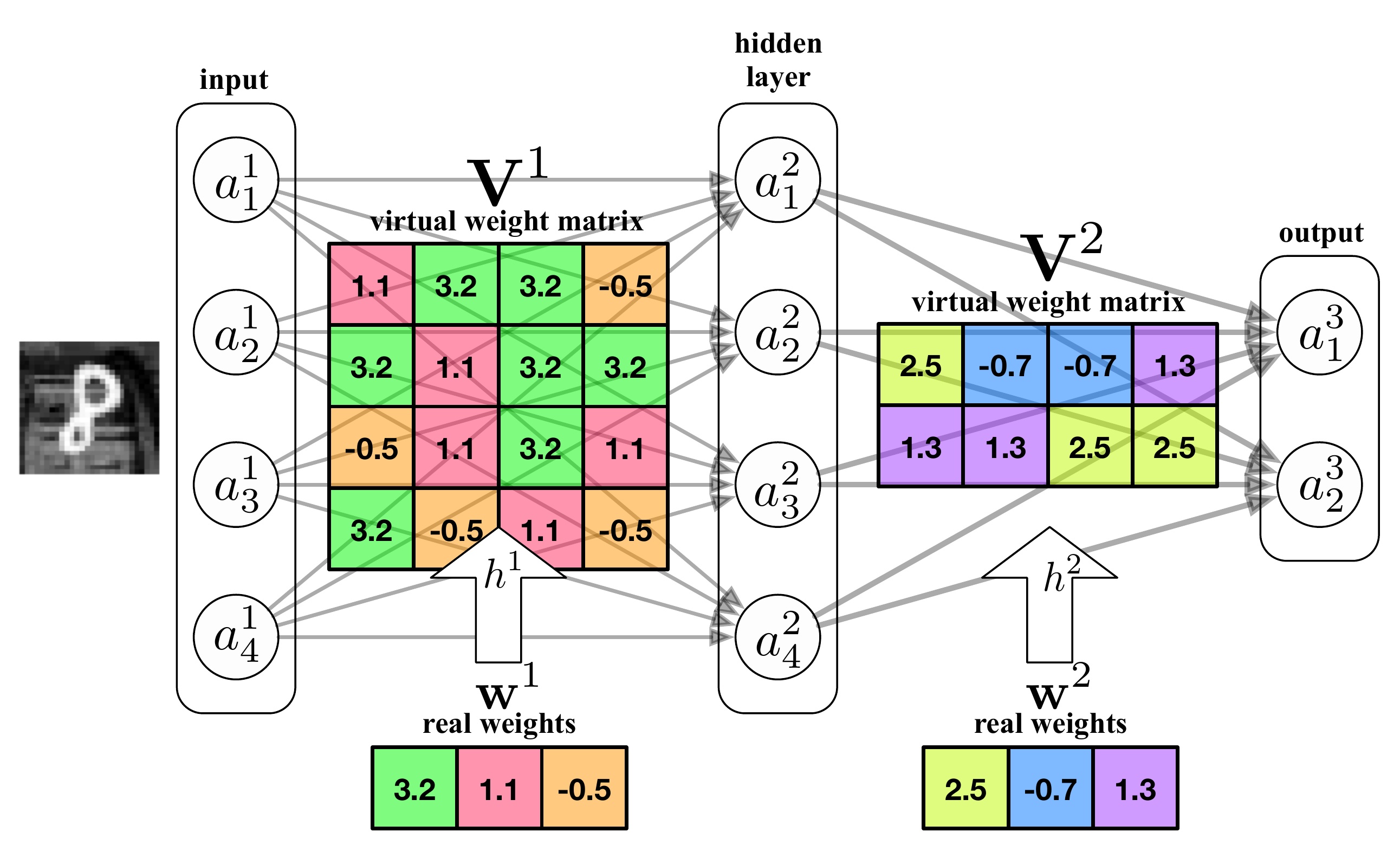}
	}
    \caption{An illustration of a neural network with random weight sharing under compression factor $\frac{1}{4}$.  The $16\!+\!9\!=\!24$ virtual weights are compressed into $6$ real weights. The colors represent matrix elements that share the same weight value.}
    \label{fig:weight_share}
\end{figure}

Figure \ref{fig:weight_share} shows a neural network with one hidden layer, four input units and two output units. Connections are randomly grouped into three categories per layer and their weights are shown in the virtual weight matrices $\bV^1$ and $\bV^2$. Connections belonging to the same color share the same weight value, which are stored in $\bw^1$ and $\bw^2$, respectively. Overall, the entire network is compressed by a factor $1/4$, \emph{i.e.} the $24$ weights stored in the virtual matrices $\bV^1$ and $\bV^2$ are reduced to only six real values in $\bw^1$ and $\bw^2$. 
On data with four input dimensions and two output dimensions, a conventional neural network with six weights would be restricted to a single (trivial) hidden unit. 

\subsection{\fullname{} (\abbrev{})}
%A na\"{i}ve implementation of random weight sharing could be trivially achieved by maintaining another matrix with the shared weight assignment for each connection.
%However, this explicit representation could require more storage overhead than the full weight matrix $\bV^\ell$.

A na\"{i}ve implementation of random weight sharing can be trivially achieved by maintaining a secondary matrix consisting of each connection's group assignment. Unfortunately, this explicit representation places an undesirable limit on potential memory savings. 
%Specifically, if each stored weight and group assignment consumes $M,N$ bits, respectively, the upper limit on the achievable compression is then $\frac{M}{N}$.
%However, this explicit representation could require more storage overhead than the full weight matrix $\bV^\ell$.

We propose to implement the random weight sharing assignments using the hashing trick.
In this way, the shared weight of each connection is determined by a hash function that requires no storage cost with the model.
Specifically, we assign to $V_{ij}^\ell$ an element of $\bw^\ell$ indexed by a hash function $\hij$, as follows:
\begin{equation}
	V_{ij}^\ell=w_{\hij}^{\ell},
	\label{eq.weight_sharing_noxi}
\end{equation}
where the (approximately uniform) hash function $\hh{\cdot}{\cdot}$ maps a key $(i,j)$ to a natural number within $\{1,\dots,K^\ell\}$. In the example of Figure~\ref{fig:weight_share}, $\hhl{1}{2}{1}\!=\!1$ and therefore \mbox{$V^1_{2,1}\!=\!w^1\!=\!3.2$}. 
For our experiments we use the open-source implementation \emph{xxHash}.\footnote{\url{https://code.google.com/p/xxhash/}}

% Note that hash function $\hh{\cdot}{\cdot}$ is characterized by the layer id $\ell$ and two neuron ids $i$ and $j$.
% While we maintain different hash \emph{tables} $\bw^\ell$ for each layer of weights, following the approach suggested by \citet{weinberger09feature}, all of these hash functions could be easily implemented by a single hash function with different keys.
% For example, $\hij$ could be implemented with a single hash function over triples $h(i,j,k)$. This would result in weight-sharing across the entire network. 

%% file: content/featurehashing.tex
%!TEX root=../hashnn_main.tex

\subsection{Feature hashing versus weight sharing}

%In the view of weight sharing, $V_{ij}$ is assigned a value from $\bw$ corresponding to the hashed index $\hijnol$ as in Eq. \eqref{eq.weight_sharing_noxi}.
This section focuses on a single layer throughout and to simplify notation we will drop the super-scripts $\ell$. We will denote the input activation as $\ba\!=\!\ba^\ell\!\in\!{\cal R}^m$ of dimensionality $m\!=\!n^\ell$. We denote the output as $\bz\!=\!\bz^{\ell+1}\!\in\!{\cal R}^{n}$ with dimensionality $n\!=\!n^{\ell+1}$. 

To facilitate weight sharing within a feed forward neural network, 
we can simply substitute Eq. \eqref{eq.weight_sharing_noxi} into Eq. \eqref{eq.output_std}:
\begin{equation}
	z_i = \sum_{j=1}^{m} V_{ij} a_j = \sum_{j=1}^{m} w_{\hijnol}  a_j.
	\label{eq.hash_output_noxi}
\end{equation}
Alternatively and more in line with previous work \cite{weinberger09feature}, we may interpret \abbrev{} in terms of feature hashing.
To compute  $z_i$, we first hash the activations from the previous layer, $\ba$, with the hash mapping function $\phi_i (\cdot)\colon\!\mR^{m}\rightarrow \mR^{K}$.
We then compute the inner product between the hashed representation $\phi_i (\ba)$ and the parameter vector $\bw$,
\begin{equation}
	z_i = \bw^\top \phi_i (\ba).
	\label{eq.fhash_output}
\end{equation}
% where $\phi_i (\ba)$ is the resulting vector of hashing input activations $\ba$.
Both $\bw$ and $\phi_i (\ba)$ are $K$-dimensional, where $K$ is the number of hash buckets in this layer. The hash mapping function $\phi_i$ is defined as follows. The $k^{th}$ element of $ \phi_i (\ba)$, \emph{i.e.} $[\phi_{i}(\ba)]_k$, is the sum of variables hashed into bucket $k$:
\begin{equation}
	[\phi_{i} (\ba)]_k = \sum_{j:\hijnol=k} a_j.
	\label{eq.myhash_noxi}
\end{equation}
% where $\hijnol:\mathbb{N}\rightarrow \{1,\cdots,K\}$ and $\xi_i:\mathbb{N}\rightarrow \{\pm 1\}$ are independent hash functions.
% We show that Eq. \eqref{eq.fhash_output} is equivalent to Eq. \eqref{eq.hash_output_noxi}.
Starting from Eq. \eqref{eq.fhash_output}, we show that the two interpretations (Eq. \eqref{eq.hash_output_noxi} and \eqref{eq.fhash_output}) are equivalent:
% \begin{eqnarray}
% 	z_i &=&  \sum_{k=1}^{K} w_k \phi_{i} (\ba) \label{eq.same_output_0} \\
% 	&=&  \sum_{k=1}^{K} w_k \sum_{j:\hijnol=k}  a_j \label{eq.same_output_1} \\
% 	&=&  \sum_{k=1}^{K} \sum_{j=1}^{n} w_k \delta_{[\hijnol=k]} a_j \label{eq.same_output_2} \\
% 	&=&  \sum_{j=1}^{n} \sum_{k=1}^{K} w_k \delta_{[\hijnol=k]} a_j \label{eq.same_output_3} \\
% 	&=& \sum_j^{n} w_{\hijnol}  a_j \label{eq.same_output_4}
% \end{eqnarray}
\begin{eqnarray}
	z_i &=&  \sum_{k=1}^{K} w_k \left[\phi_{i} (\ba)\right]_k =\sum_{k=1}^{K} w_k \sum_{j:\hijnol=k}  a_j \nonumber\\
	&=&   \sum_{j=1}^{m}\sum_{k=1}^{K} w_k  a_j \delta_{[\hijnol=k]}\nonumber \\
	&=& \sum_{j=1}^{m} w_{\hijnol}  a_j. \nonumber
\end{eqnarray}
%where Eq. \eqref{eq.same_output_1} is obtained by substituting Eq. \eqref{eq.myhash_noxi} into Eq. \eqref{eq.same_output_0}. Then in Eq. \eqref{eq.same_output_2} we make the predicate $\hijnol=k$ in the second summation explicit by a delta function $\delta$ where $\delta_{\hijnol=k}=1$ when $\hijnol=k$ and 0 otherwise. Eq. \eqref{eq.same_output_3} flips the summation of Eq. \eqref{eq.same_output_2}. 
The final term is equivalent to Eq. \eqref{eq.hash_output_noxi}. 

% We can see that Eq. \eqref{eq.same_output_4} is identical to Eq. \eqref{eq.hash_output_noxi}.

\paragraph{Sign factor.} 
With this equivalence between random weight sharing and feature hashing on input activations, \abbrev{} inherit several beneficial properties of the feature hashing.
\citet{weinberger09feature} introduce an additional sign factor $\xi(i,j)$ to remove the bias of hashed inner-products due to collisions. 
For the same reasons we multiply \eqref{eq.weight_sharing_noxi} by the sign factor $\xi(i,j)$ for parameterizing $\bV$~\cite{weinberger09feature}:
\begin{equation}
	V_{ij}=w_{\hijnol} \xi(i,j),
	\label{eq.weight_sharing}
\end{equation}
where $\xi(i,j)\colon\!\mathbb{N}\rightarrow {\pm 1}$ is a second hash function independent of $h$.
Incorporating $\xi(i,j)$ to feature hashing and weight sharing does not change the equivalence between them as the proof in the previous section still holds with the sign term (details omitted for improved readability).

%With $\xi$, \abbrev{} enjoy the property of approximate orthogonality and the inner products become unbiased in expectation~\cite{weinberger09feature}. 

% \paragraph{Unbiased approximation.}
% Let $\ba_1,\dots,\ba_N$ denote the inputs to the first layer (\emph{i.e.} $\ba_i$ is the feature vector of the $i^{th}$ training input). Let the  $\mathbf{g}_i$ denote the gradient for the $i^{th}$ row of the \emph{actual} (not virtual) version of $\bV$. We must have $\mathbf{g}_i=\sum_{j=1}^N\alpha_j\ba_j$ for some weights $\alpha_j$. The hashed gradient 

% , \emph{i.e.} $\phi_i(\ba)^\top \phi_j(\ba')\approx 0$, for $i\neq j$. 
%  $\bw=\sum_{i=1}^N$

% leading to an unbiased approximation to the original fully connected network, \emph{i.e.} $\bw^{\top} \phi_i(\ba)\approx \bV_i^{\top} \ba$.

\paragraph{Sparsity.} 
As pointed out in \citet{hashKernelShi:2009} and \citet{weinberger09feature}, feature hashing is most effective on sparse feature vectors since the number of hash collisions is minimized. We can encourage this effect in the hidden layers with sparsity inducing transition functions, \emph{e.g.} rectified linear units (ReLU)~\cite{glorot2011deep} or through specialized regularization~\cite{icml2014c2_cheng14,boureau2008sparse}. In our implementation, we use ReLU transition functions throughout, as they have also been shown to often result in superior generalization performance in addition to their sparsity inducing properties~\cite{glorot2011deep}. 

\paragraph{Alternative neural network architectures.}
While this work focuses on general, fully connected feed forward neural networks, the technique of \abbrev{} could naturally be extended to other kinds of neural networks, such as recurrent neural networks~\cite{pineda1987generalization} or others~\cite{bishop1995neural}. It can also be used in conjunction with other approaches for neural network compression. All weights can be stored with low bit precision~\cite{courbariaux2015low,gupta2015deep}, edges could be removed~\cite{cirecsan2011high} and \abbrev{} can be trained on the outputs of larger networks \cite{Caruana2014} --- yielding further reductions in memory requirements.  

%\citet{denil2013predicting} point out that especially convolutional neural networks have strong correlation between their weights, due to local smoothness of input pixels. 

 % \begin{figure*}[t]
 %     \centerline{
 %      \includegraphics[width=0.8\textwidth]{figures/datasets.pdf}
 %      }
 %      \caption{Samples from various image datasets}
 %    \label{fig:dataset}
 % \end{figure*}

%% file: content/backprop.tex
%!TEX root=../hashnn_main.tex

\subsection{Training \abbrev{}}
Training \abbrev{} is equivalent to training a standard neural network with equality constraints for weight sharing. Here, we show how to (a) compute the output of a hash layer during the feed-forward phase, (b) propagate gradients from the output layer back to input layer, and (c) compute the gradient over the shared weights $\bw^\ell$ during the back propagation phase. We use dedicated hash functions between layers $\ell$ and $\ell+1$, and denote them as $h^\ell$ and $\xi^\ell$. 

\paragraph{Output.} Adding the hash functions $h^{\ell}(\cdot,\cdot)$ and  $\xi^{\ell}(\cdot)$ and the weight vectors $\bw^\ell$ into the feed forward update~\eqref{eq.output_std} results in the following forward propagation rule:
\begin{equation}
	a^{\ell+1}_i =  f\left(\sum_j^{n^{\ell}} w_{\hij}^{\ell} \xij a_j^\ell\right).
	\label{eq.hash_output}
\end{equation}

\paragraph{Error term.} 
Let $\cal L$ denote the loss function for training the neural network, \emph{e.g.} cross entropy or the quadratic loss~\cite{bishop1995neural}. 
Further, let $\delta_j^{\ell}$ denote the gradient of $\cal L$ over activation $j$ in layer $\ell$, also known as the error term. 
Without shared weights, the error term can be expressed as $\delta_j^{\ell} = \left( \sum_{i=1}^{n^{\ell+1}} V^\ell_{ij}  \delta^{\ell+1}_{i} \right) f^\prime(z^{\ell}_j)$, where $f'(\cdot)$ represents the first derivative of the transition function $f(\cdot)$. If we substitute Eq. \eqref{eq.weight_sharing} into the error term we obtain:
\begin{equation}
	\delta_j^{\ell} = \left( \sum_{i=1}^{n^{\ell+1}} \xx{i}{j}  w^{\ell}_{\hh{i}{j}}  \delta^{\ell+1}_{i} \right) f^\prime(z^{\ell}_j).
	\label{eq.error_term}
\end{equation}

\paragraph{Gradient over parameters.}
To compute the gradient of ${\cal L}$ with respect to a weight $w_k^\ell$ we need the two gradients, 
\begin{equation}
	\gradL{V^\ell_{ij}} = a_j^\ell \delta_i^{\ell+1} \textrm{ and } 
\grad{V^\ell_{ij}}{w_k^\ell}=\xij\delta_{\hij=k}.
	\label{eq.normal_gradient}
\end{equation}
Here, the first gradient is the standard gradient of a (virtual) weight with respect to an activation unit and the second gradient ties the virtual weight matrix to the actual weights through the hashed map. 
Combining these two, we obtain
\begin{eqnarray}
	\gradL{w_k^\ell} &=& \sum_{i,j} \gradL{V^\ell_{ij}} \grad{V^\ell_{ij}}{w_k^\ell} \label{eq.gradchain} \\
	 			  %&=& \sum_{i,j:\hij=k} \xij \gradL{ V^\ell_{ij} } \label{eq.grad1}\\
				  %&=& \sum_{i=1}^{n^{\ell+1}} \sum_{j:\hij=k} \xij \gradL{V^\ell_{ij} } \label{eq.grad2}\\
				  &=& \sum_{i=1}^{n^{\ell+1}} \sum_{j}  a_j^\ell \delta_i^{\ell+1} \xij\delta_{\hij=k}\label{eq.grad3}.
\end{eqnarray}

%where Eq. \eqref{eq.gradchain} is based on chain rule for computing derivatives while Eq. \eqref{eq.grad1} can be obtained by combining with Eq. \eqref{eq.weight_sharing}. Eq. \eqref{eq.grad2} is an rearrangement of Eq. \eqref{eq.grad1}, and Eq. \eqref{eq.grad3} is the direct result of substituting Eq. \eqref{eq.normal_gradient}.

%% file: content/related.tex
%!TEX root=../hashnn_main.tex

\section{Related Work}

Deep neural networks have achieved great progress on a wide variety of real-world applications, including image classification~\cite{krizhevsky2012imagenet,donahue2013decaf,sermanet2013overfeat,zeiler2014visualizing}, object detection~\cite{girshick2014rich,vinyals2014show}, image retrieval \cite{razavian2014cnn}, speech recognition \cite{hinton2012deep,graves2013speech,mohamed2011deep}, and text representation~\cite{mikolov2013distributed}.
%These recent advances have propelled deep learning to the center of popular attention.

There have been several previous attempts to reduce the complexity of neural networks under a variety of contexts.
Arguably the most popular method is the widely used convolutional neural network \cite{simard2003best}.
In the convolutional layers, the same filter is applied to every receptive field, both reducing model size and improving generalization performance.
The incorporation of pooling layers \cite{zeiler2013stochastic} can reduce the number of connections between layers in domains exhibiting locality among input features, such as images.
Autoencoders \cite{glorot2011domain} share the notion of tied weights by using the same weights for the encoder and decoder (up to transpose). 
%in which the encoder and decoder share the same weight matrix (up to a transpose).

Other methods have been proposed explicitly to reduce the number of free parameters in neural networks, but not necessarily for reducing memory overhead.
\citet{nowlan1992simplifying} introduce soft weight sharing for regularization in which the distribution of weight values is modeled as a Gaussian mixture.
The weights are clustered such that weights in the same group have similar values.
Since weight values are unknown before training, weights are clustered during training.
This approach is fundamentally different from \abbrev{} since
 % soft-weight sharing only reduces the number of free parameters.
it requires auxiliary parameters to record the group membership for every weight.
% , which in fact doesn't reduce memory consumption.
% In addition, instead of telling the network which weights should be identical, \abbrev{} randomly shares the weights among all connections and let the training adjust the neural network itself to fit this random assigned structure.

Instead of sharing weights, \citet{lecun1989optimal} introduce ``optimal brain damage'' to directly drop unimportant weights.
This approach requires auxiliary parameters for storing the sparse weights and needs retraining time to fine-tune the resulting architecture. \citet{cirecsan2011high} demonstrate in their experiments that randomly removing connections leads to superior empirical performance, which shares the same spirit of \abbrev{}.

\citet{courbariaux2015low} and \citet{gupta2015deep} learn networks with reduced numerical precision for storing model parameters (\emph{e.g.} $16$-bit fixed-point representation \cite{gupta2015deep} for a compression factor of $\frac{1}{4}$ over double-precision floating point). Experiments indicate little reduction in accuracy compared with models trained with double-precision floating point representation. These methods can be readily incorporated with \abbrev{}, potentially yielding further reduction in model storage size.

A recent study by \citet{denil2013predicting} demonstrates significant redundancy in neural network parameters by directly learning a low-rank decomposition of the weight matrix within each layer.
They demonstrate that networks composed of weights recovered from the learned decompositions are only slightly less accurate than networks with all weights as free parameters, indicating heavy over-parametrization in full weight matrices.
A follow-up work by \citet{denton2014} uses a similar technique to speed up test-time evaluation of convolutional neural networks. 
The focus of this line of work is not on reducing storage and memory overhead, but evaluation speed during test time. \abbrev{} is complementary to this research, and the two approaches could be used in combination. 

Following the line of model compression, \citet{bucilua2006model}, \citet{hinton2014} and \citet{Caruana2014} recently introduce approaches to learn a ``distilled'' model, training a more compact neural network to reproduce the output of a larger network.
Specifically, \citet{hinton2014} and \citet{Caruana2014} train a large network on the original training labels, then learn a much smaller ``distilled'' model on a weighted combination of the original labels and the (softened) softmax output of the larger model.
% whose softmax outputs on training samples are then soften by a constant named temperature.
% Next, a much smaller distilled model is trained on the same training dataset with two objectives in a weighted sum fashion: the cross entropy with true labels and the cross entropy with soften targets, respectively.
The authors show that the distilled model has better generalization ability than a model trained on just the labels.
% , which obtains superior results on a bunch of experiments.
% Following the settings in \citet{hinton2014}, we were able to reproduce the results when the distilled model is trained without dropout.
% However, in our settings, with dropout, we find that the distilled model is typically outperformed by training an equivalently-sized network on the original labels, especially when the distilled model is much smaller than the large network. We therefore omit these results in the experimental Section~\ref{sec:exp} and include results from the better-performing neural networks with dropout instead (trained on the true labels).
% So in the result section, we only train distilled model on the real targets.
In our experimental results, we show that our approach is complementary by learning \abbrev{} with soft targets.
\citet{rippel2014learning} propose a novel dropout method, nested dropout, to give an order of importance for hidden neurons.
Hypothetically, less important hidden neurons could be removed after training, a method orthogonal to \abbrev{}.

\citet{ganchev:08} are among the first to recognize the need to reduce the size of natural language processing models to accommodate mobile platform with limited memory and computing power.
They propose \emph{random feature mixing} to group features at random based on a hash function, which dramatically reduces both the number of features and the number of parameters.
With the help of feature hashing~\cite{weinberger09feature}, \textit{Vowpal Wabbit}, a large-scale learning system, is able to scale to terafeature datasets \cite{agarwal2014reliable}.

%% file: content/experiment.tex
 \begin{figure*}[t]
 \vspace{-1ex}
     \centerline{
      \includegraphics[width=0.40\textwidth]{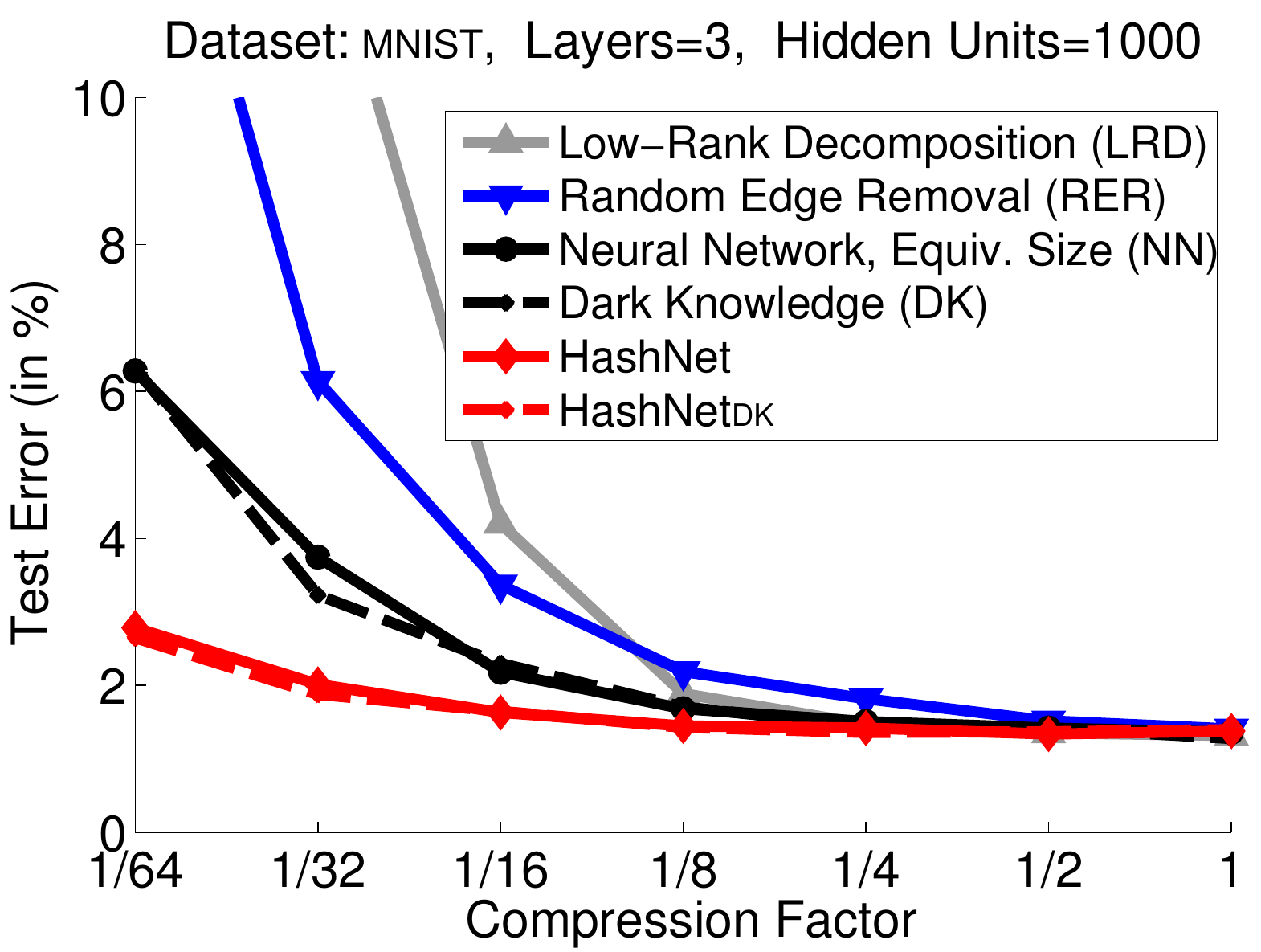}
	  ~~~~~~~~
      \includegraphics[width=0.40\textwidth]{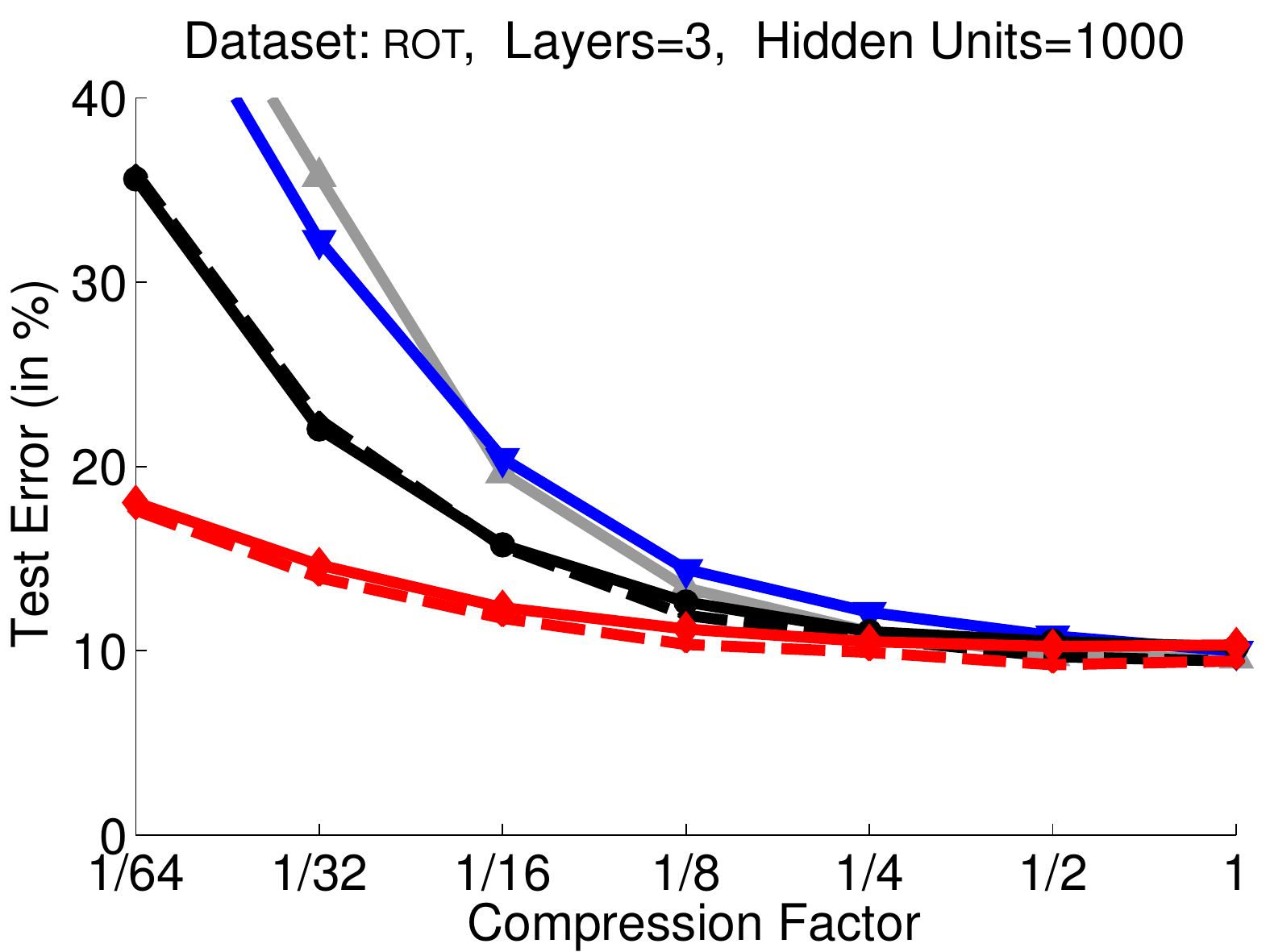}
      }            
      \vspace{-1.5ex}
    \caption{Test error rates under varying compression factors with $3$-layer networks on {\sc{mnist}} (\emph{left}) and {\sc{rot}} (\emph{right}).
}
    \label{fig:compressl3}
 \end{figure*}
\section{Experimental Results}
\label{sec:exp}

 \begin{figure*}[t]
\vspace{-0.5ex}
     \centerline{
      \includegraphics[width=0.40\textwidth]{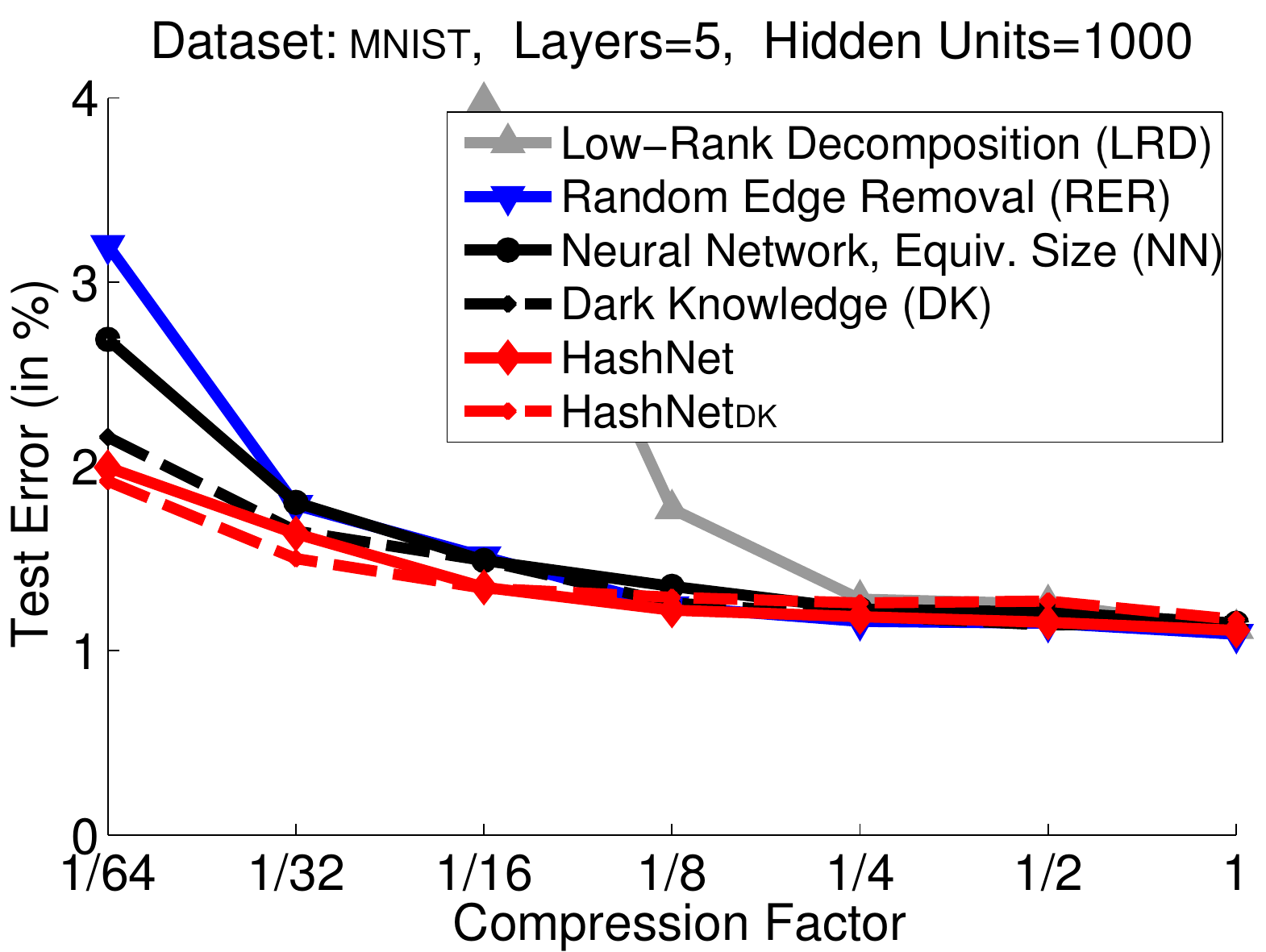}
	  ~~~~~~~~
      \includegraphics[width=0.40\textwidth]{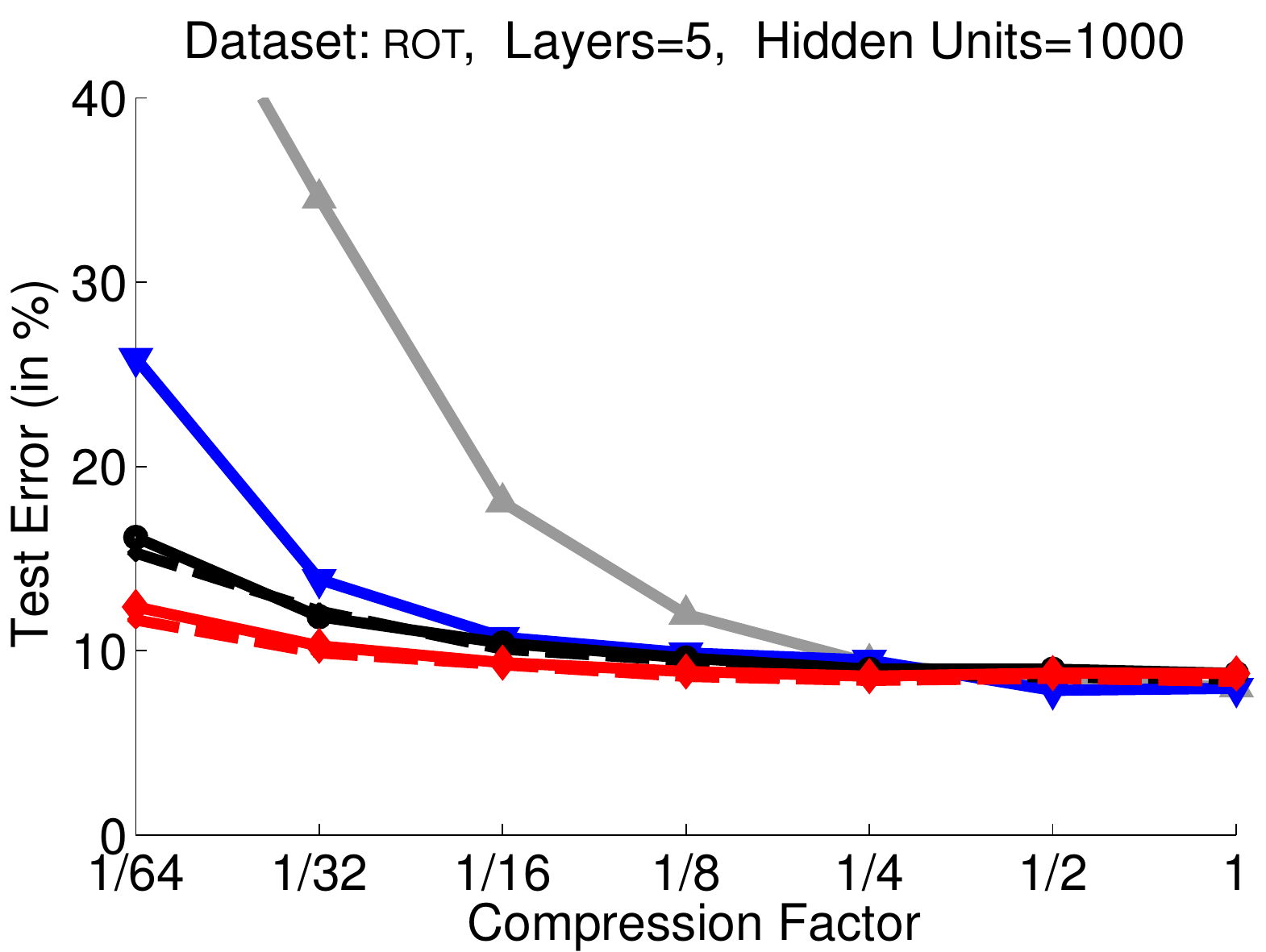}
      }\vspace{-1.5ex}
          \caption{Test error rates under varying compression factors with $5$-layer networks on {\sc{mnist}} (\emph{left}) and {\sc{rot}} (\emph{right}).}
    \label{fig:compressl5}
 \end{figure*}
 
\input content/resulttables.tex
\begin{figure*}[t]
\vspace{-0.5ex}
    \centerline{
    \includegraphics[width=0.40\textwidth]{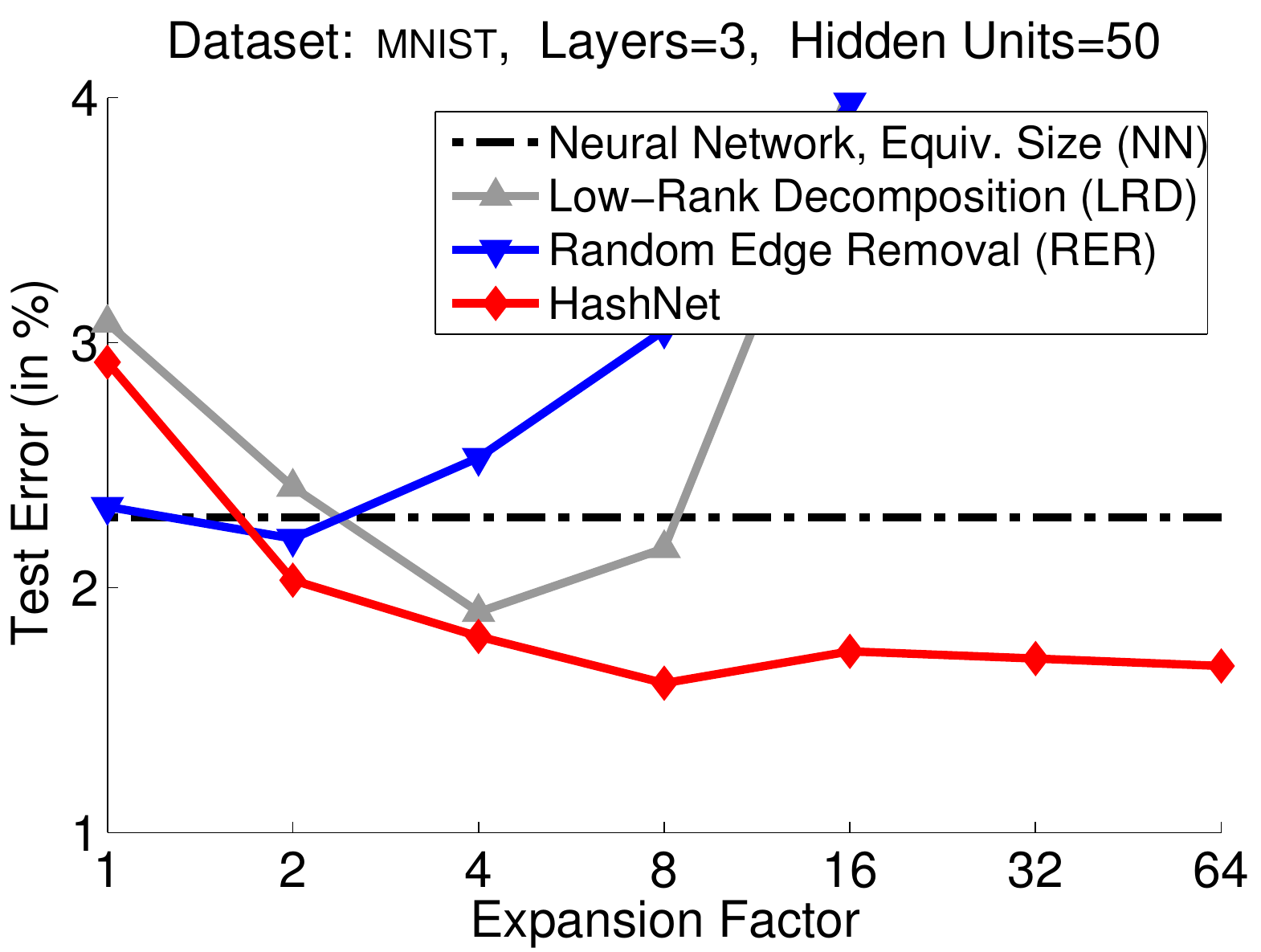}
	~~~~~~~~
    \includegraphics[width=0.40\textwidth]{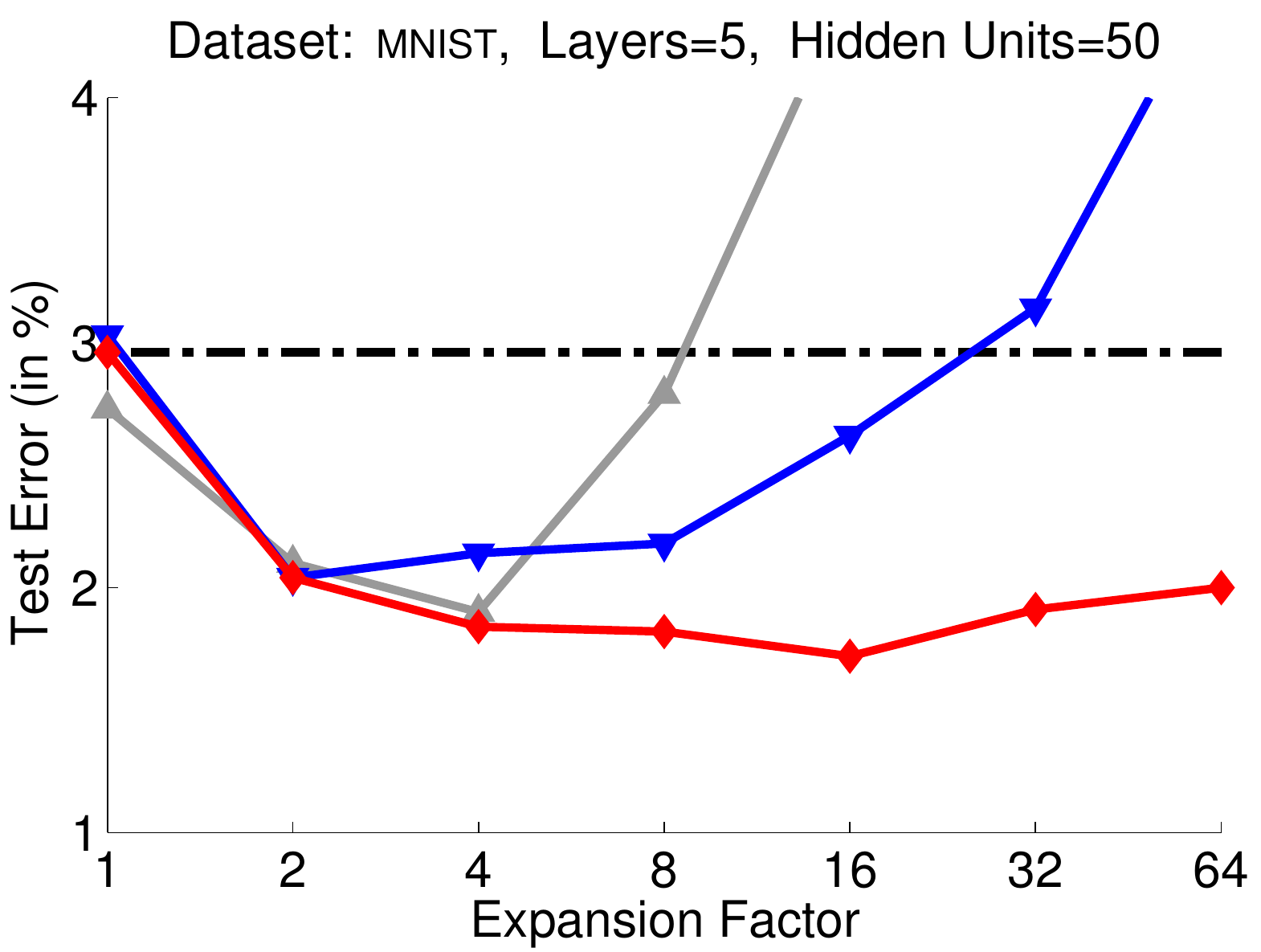}
  }\vspace{-1ex}
    \caption{Test error rates with fixed storage but varying expansion factors on {\sc{mnist}} with 3 layers (\emph{left}) and 5 layers (\emph{right}). }
    \label{fig:fixed_storage}
\end{figure*}

We conduct extensive experiments to evaluate \abbrev{} on eight benchmark datasets.
For full reproducibility, our code is available at {\small\url{http://www.weinbergerweb.com}}.

%\subsection{Setup}

\paragraph{Datasets.}
Datasets consist of the original {\sc{mnist}} handwritten digit dataset, along with four challenging variants \cite{larochelle2007empirical}.
Each variation amends the original through digit rotation ({\sc{rot}}), background superimposition ({\sc{bg-rand}} and {\sc{bg-img}}), or a combination thereof ({\sc{bg-img-rot}}).
In addition, we include two binary image classification datasets: {\sc{convex}} and {\sc{rect}} \cite{larochelle2007empirical}.
All data sets have pre-specified training and testing splits.
Original {\sc{mnist}} has splits of sizes $n\!=\!60000$ (training) and $n\!=\!10000$ (testing).
Both {\sc{convex}} and {\sc{rect}} and as well as each {\sc{mnist}} variation set has $n\!=\!12000$ (training) and $n\!=\!50000$ (testing).

% Further details regarding the mentioned datasets can be found in [Larochelle, CITE].

%Our first dataset is the original MNIST 
%dataset (MNIST) for recognizing images of handwritten
%digits. Besides, we also experiment on its several variants \cite{rifai2011contractive}
% which are more challenging modification
%to the MNIST dataset. These include images of rotated digits
%(rot), images superimposed onto random (bg-rand) or image
%background (bg-img) and the combination of rotated
%digits with image background (bg-img-rot). We also test all methods
%on three shape classification tasks (convex, rect,
%rect-img).

\paragraph{Baselines and method.}
We compare \abbrev{} with several existing techniques for size-constrained, feed-forward neural networks.
\emph{Random Edge Removal} (RER) \cite{cirecsan2011high} reduces the total number of model parameters by randomly removing weights prior to training.
\emph{Low-Rank Decomposition} (LRD) \cite{denil2013predicting} decomposes the weight matrix into two low-rank matrices. One of these component matrices is fixed while the other is learned. 
Elements of the fixed matrix are generated according to a zero-mean Gaussian distribution with standard deviation $\frac{1}{\sqrt{n^\ell}}$ with $n^\ell$ inputs to the layer.

Each model is compared against a standard neural network with an equivalent number of stored parameters, \emph{Neural Network (Equivalent-Size)} (NN).
For example, for a network with a single hidden layer of $1000$ units and a storage compression factor of $\frac{1}{10}$, we adopt a size-equivalent baseline with a single hidden layer of $100$ units.
For deeper networks, all hidden layers are shrunk at the same rate until the number of stored parameters equals the target size.
In a similar manner, we examine \emph{Dark Knowledge} (DK)~\cite{hinton2014,Caruana2014} by training a distilled model to optimize the cross entropy with both the original labels and soft targets generated by the corresponding full neural network (compression factor $1$).
The distilled model structure is chosen to be same as the ``equivalent-sized'' network (NN) at the corresponding compression rate.

Finally, we examine our method under two settings: learning hashed weights with the original training labels (\abbrevtable{}) and with combined labels and DK soft targets (\abbrevtabledk{}).
% For space limit, we only present the results of using NN and \abbrev{} as the distilled models, under the name of NN$_{DK}$ and \abbrev{}$_{DK}$, respectively.
% \kilian{Need description of this here.}
In all cases, memory and storage consumption is defined strictly in terms of free parameters.
As such, we count the fixed low rank matrix in the Low-Rank Decomposition method as taking no memory or storage (providing this baseline a slight advantage).

%
%generates a new network structure by dropping connections at random from fully-connected network. Note that the new structure is fixed throughout the training. Low-rank Decomposition (decompG) \cite{denil2013predicting} decomposes the weight matrix with two low-rank matrices $U$ and $V$, where $V$ is the free parameters to be learned. In our experiment, we use Gaussian distribution to generate $U$. Another simple but effective baseline is the standard neural network with reduced hidden units. For example, if we want to compress a 1-hidden-layer neural network with 1000 hidden units by 10 times, we could just reduce the number of hidden units from 1000 to 100. Surprisingly, this is a quite competitive method and few papers have ever compared with this baseline.
%
%For all aforementioned baseline methods, we only count their free parameters for memory consumption. For example, we consider $U$ matrix taking no memory even though they might take massive amount of memory in practice.

\paragraph{Experimental setting.}

\abbrev{} and all accompanying baselines were implemented using Torch7 \cite{collobert2011torch7} and run on NVIDIA GTX TITAN graphics cards with $2688$ cores and $6$GB of global memory. We use $32$ bit precision throughout but note that the compression rates of all methods may be improved with lower precision~\cite{courbariaux2015low,gupta2015deep}.
We verify all implementations by numerical gradient checking.
Models are trained via stochastic gradient descent (mini-batch size of $50$) with dropout and momentum.
ReLU is adopted as the activation function for all models.
%  selected using a combination of Bayesian Optimization~\cite{snoek2012practical,gardner2014bayesian} and hand-tuning.
% All the reported results are made on the pre-defined test sets for each problem.
Hyperparameters are selected for all algorithms with Bayesian optimization~\citep{snoek2012practical} and hand tuning on $20\%$ validation splits of the training sets. We use the open source Bayesian Optimization MATLAB implementation \href{https://bitbucket.org/mlcircus/bayesopt.m}{``bayesopt.m''} from~\citet{gardner2014bayesian}.\footnote{\url{http://tinyurl.com/bayesopt}}

%All models including \abbrev{} are implemented with Torch7 \cite{collobert2011torch7} and executed in GPU. In addition, all these models are trained with dropout and \textit{min-batch} gradient descent with momentum. The hyperparameters are tuned on validation set with both Bayesian Optimization~\cite{snoek2012practical,gardner2014bayesian} and manually adjust. All the reported results are evaluated on testing set.
%All the experiments were performed on an off-the-shelve desktop with a NVIDIA GTX TITAN graphics card attached which contains 2688 cores and 6 GB of global memory.

\paragraph{Results with varying compression.}
Figures \ref{fig:compressl3} and \ref{fig:compressl5} show the performance of all methods on {\sc{mnist}} and the {\sc{rot}} variant with different compression factors on $3$-layer ($1$ hidden layer) and $5$-layer ($3$ hidden layers) neural networks, respectively.
Each hidden layer contains $1000$ hidden units.
% For example, the structure for $5$-layer network for MNIST in our experiment is $[784\ 1000\ 1000\ 1000\ 10]$.
The $x$-axis in each figure denotes the fractional compression factor.
For \abbrev{} and the low rank decomposition and random edge removal compression baselines, this means we fix the number of hidden units ($n^\ell$) and vary the storage budget ($K^\ell$) for the weights ($\bw^{\ell}$).

We make several observations:
The accuracy of \abbrevtable{} and \abbrevtabledk{} outperforms all other baseline methods, especially in the most interesting case when the compression factor is small (\emph{i.e.} very small models).
 %Without compression, the accuracy of the various methods consistently converges to a common performance level.
Both compression baseline algorithms, low rank decomposition %\citet{denil2013predicting}
and random edge removal, 
tend to not outperform a standard neural network with fewer hidden nodes (black line), trained with dropout.
%The low-rank method slightly outperforms  with no compression; however, for compression factors greater than $\frac{1}{8}$, its performance quickly falls off especially with the deeper $5$-layer networks.
% One possible explanation is the accumulated random input projections   that the input at each layer is random projected to a much smaller feature space where the value of each random feature is the sum of many random variables and it tends to be the same across different random features. This makes the neural network hard to be trained.
For smaller compression factors, random edge removal likely suffers due to a significant number of nodes being entirely disconnected from neighboring layers.
The size-matched NN is consistently the best performing baseline, however its test error is significantly higher than that of \abbrevtable{} especially at small compression rates.
The use of Dark Knowledge training improves the performance of \abbrev{} and the standard neural network.
% when compared with \abbrev{} and other baselines.
Of all methods, only \abbrevtable{} and \abbrevtabledk{} maintain performance for small compression factors.

% \noindent
% Observations:
% \begin{itemize}
% \item[-]
% \item[-]
% \item[-] The low-rank method's lackluster performance can, in some senses, be excused as reflecting the fact the goal of the method is to reduce the number of free parameters (not the total number of parameters)
% \item[-]
% \item[-]
% \item[-]
% \end{itemize}

For completeness, we show the performance of all methods on all eight datasets in Table \ref{tab:table_1} for compression factor $\frac{1}{8}$ and Table~\ref{tab:table_2} for compression factor $\frac{1}{64}$. 
%The respective models store only  1 \emph{byte} (Table~\ref{tab:table_1}) or 1 \emph{bit} (Table~\ref{tab:table_2})  per parameter. 
\abbrevtable{} and \abbrevtabledk{} outperform other baselines in most cases, especially when the compression factor is very small (Table~\ref{tab:table_2}). With a compression factor of $\frac{1}{64}$ on average only $0.5$ \emph{bits} of information are stored per (virtual) parameter. 

\paragraph{Results with fixed storage.}
We also experiment with the setting where the model size is fixed and the virtual network architecture is ``inflated''. 
Essentially we are fixing $K^\ell$ (the number of ``real'' weights in $\bw^\ell$), and vary the number of hidden nodes ($n^{\ell}$).
An expansion factor of $1$ denotes the case where every virtual weight has a corresponding ``real'' weight, $(n^{\ell}+1)n^{\ell+1}\!=\!K^\ell$.
Figure \ref{fig:fixed_storage} shows the test error rate under various expansion rates of a network with one hidden layer (\emph{left}) and three hidden layers (\emph{right}).
In both scenarios we fix the number of real weights to the size of a standard fully-connected neural network with $50$ hidden units in each hidden layer whose test error is shown by the black dashed line. %$K^\ell\!=\!50$.
%The red dashed line shows the test error rate of a neural network with the same number of free parameters ($50$). 
%The red dashed line shows the test error rate of this neural network. 

With no expansion (at expansion rate $1$), different compression methods perform differently. At this point edge removal is identical to a standard neural network and matches its results. If no expansion is performed, the \abbrevtable{} performance suffers from collisions at no benefit. Similarly the low-rank method still randomly projects each layer to a random feature space with same dimensionality.

For expansion rates greater $1$, all methods improve over the fixed-sized neural network. There is a general trend that more expansion decreases the test error until a ``sweet-spot'' after which additional expansion tends to hurt. 
The test error of the \abbrevtable{} neural network decreases substantially through the introduction of more ``virtual'' hidden nodes, despite that no additional parameters are added. In the case of the 5-layer neural network (right) this trend is maintained to an expansion factor of $16\times$, resulting in $800$ ``virtual'' nodes. %, with  $640000$ virtual weights which are hashed into mere $50$ real weights. 
One could hypothetically increase $n^\ell$ arbitrarily for \abbrevtable{}, however, in the limit, too many hash collisions would result in increasingly similar gradient updates for all weights in $\bw$. 

The benefit from expanding a network cannot continue forever. In the \emph{random edge removal} the network will become very sparsely connected; the low-rank decomposition approach will eventually lead to a decomposition into rank-$1$ matrices.
\abbrevtable{} also respects this trend, but is much less sensitive when the expansion goes up. Best results are achieved when networks are inflated by a factor $8\!-\!16\times$.

%% file: content/resulttables.tex
%!TEX root=../hashnn_main.tex

\begin{table*}[t]
\centering      
\resizebox{\linewidth}{!}{                                                                                   
\begin{tabular}{r||c|c|c|c|c|c||c|c|c|c|c|c}
% \hline
\multicolumn{1}{c||}{} & \multicolumn{6}{c||}{3 Layers} &\multicolumn{6}{c}{5 Layers} \\
% \hline 
 & RER & LRD & NN & DK & \abbrevtable{} & \abbrevtabledk{} & RER & LRD & NN & DK & \abbrevtable{} & \abbrevtabledk{} \\  
\hline
{\sc mnist} &$ 2.19 $&$ 1.89 $&$ 1.69 $&$ 1.71 $&$ 1.45 $&$ \boldBlue{1.43} $&$ 1.24 $&$ 1.77 $&$ 1.35 $&$ 1.26 $&$ \boldBlue{1.22} $&$ 1.29 $\\                 
% \hline
{\sc basic} &$ 3.29 $&$ 3.73 $&$ 3.19 $&$ 3.18 $&$ 2.91 $&$ \boldBlue{2.89} $&$ 2.87 $&$ 3.54 $&$ 2.73 $&$ 2.87 $&$ \boldBlue{2.62} $&$ 2.85 $\\                 
% \hline
{\sc rot} &$ 14.42 $&$ 13.41 $&$ 12.65 $&$ 11.93 $&$ 11.17 $&$ \boldBlue{10.34} $&$ 9.89 $&$ 11.98 $&$ 9.61 $&$ 9.46 $&$ 8.87 $&$ \boldBlue{8.61} $\\            
% \hline
{\sc bg-rand} &$ 18.16 $&$ 45.12 $&$ 13.00 $&$ 12.41 $&$ 13.38 $&$ \boldBlue{12.27} $&$ 11.31 $&$ 45.02 $&$ 11.19 $&$ 10.91 $&$ \boldBlue{10.76} $&$ 10.96 $\\   
% \hline
{\sc bg-img} &$ 24.18 $&$ 38.83 $&$ 20.93 $&$ 19.31 $&$ 22.57 $&$ \boldBlue{18.92} $&$ 19.81 $&$ 35.06 $&$ 19.33 $&$ 18.94 $&$ 19.07 $&$ \boldBlue{18.49} $\\    
% \hline
{\sc bg-img-rot} &$ 59.29 $&$ 67.00 $&$ 52.90 $&$ 53.01 $&$ 51.96 $&$ \boldBlue{50.05} $&$ \boldBlue{45.67} $&$ 64.28 $&$ 48.47 $&$ 48.22 $&$ 46.67 $&$ 46.78 $\\
% \hline
{\sc rect} &$ 27.32 $&$ 32.73 $&$ 23.91 $&$ 24.74 $&$ 27.06 $&$ \boldBlue{22.93} $&$ 27.13 $&$ 35.79 $&$ 24.58 $&$ \boldBlue{23.86} $&$ 29.58 $&$ 25.99 $\\      
% \hline
{\sc convex} &$ 3.69 $&$ 4.56 $&$ 4.24 $&$ 3.07 $&$ 3.23 $&$ \boldBlue{2.96} $&$ 3.92 $&$ 7.09 $&$ 3.43 $&$ 2.37 $&$ 3.92 $&$ \boldBlue{2.36} $\\                
% \hline
\end{tabular}}                                                                                    
\caption{Test error rates (in \%) with a compression factor of \(\frac{1}{8}\) across all data sets. Best results are printed in \boldBlue{blue}.}                                               
\label{tab:table_1}                                                                         
\end{table*}

\begin{table*}[t]
\centering    
\resizebox{\linewidth}{!}{
\begin{tabular}{r||c|c|c|c|c|c||c|c|c|c|c|c}
% \hline
\multicolumn{1}{c||}{} & \multicolumn{6}{c||}{3 Layers} &\multicolumn{6}{c}{5 Layers} \\
% \hline 
 & RER & LRD & NN & DK & \abbrevtable{} & \abbrevtabledk{} & RER & LRD & NN & DK & \abbrevtable{} & \abbrevtabledk{} \\  
\hline
{\sc mnist} &$ 15.03 $&$ 28.99 $&$ 6.28 $&$ 6.32 $&$ 2.79 $&$ \boldBlue{2.65} $&$ 3.20 $&$ 28.11 $&$ 2.69 $&$ 2.16 $&$ 1.99 $&$ \boldBlue{1.92} $\\              
% \hline  
{\sc basic} &$ 13.95 $&$ 26.95 $&$ 7.67 $&$ 8.44 $&$ 4.17 $&$ \boldBlue{3.79} $&$ 5.31 $&$ 27.21 $&$ 4.55 $&$ 4.07 $&$ 3.49 $&$ \boldBlue{3.19} $\\              
% \hline    
{\sc rot} &$ 49.20 $&$ 52.18 $&$ 35.60 $&$ 35.94 $&$ 18.04 $&$ \boldBlue{17.62} $&$ 25.87 $&$ 52.03 $&$ 16.16 $&$ 15.30 $&$ 12.38 $&$ \boldBlue{11.67} $\\       
% \hline      
{\sc bg-rand} &$ 44.90 $&$ 76.21 $&$ 43.04 $&$ 53.05 $&$ 21.50 $&$ \boldBlue{20.32} $&$ 90.28 $&$ 76.21 $&$ 16.60 $&$ 14.57 $&$ 16.37 $&$ \boldBlue{13.76} $\\   
% \hline        
{\sc bg-img} &$ 44.34 $&$ 71.27 $&$ 32.64 $&$ 41.75 $&$ 26.41 $&$ \boldBlue{26.17} $&$ 55.76 $&$ 70.85 $&$ 22.77 $&$ 23.59 $&$ 22.22 $&$ \boldBlue{20.01} $\\    
% \hline          
{\sc bg-img-rot} &$ 73.17 $&$ 80.63 $&$ 79.03 $&$ 77.40 $&$ 59.20 $&$ \boldBlue{58.25} $&$ 88.88 $&$ 80.93 $&$ 53.18 $&$ 53.19 $&$ \boldBlue{51.93} $&$ 54.51 $\\
% \hline            
{\sc rect} &$ 37.22 $&$ 39.93 $&$ 34.37 $&$ 31.85 $&$ 31.77 $&$ \boldBlue{30.43} $&$ 50.00 $&$ 39.65 $&$ 29.76 $&$ \boldBlue{26.95} $&$ 29.70 $&$ 32.04 $\\      
% \hline              
{\sc convex} &$ 18.23 $&$ 23.67 $&$ 5.68 $&$ 5.78 $&$ 3.67 $&$ \boldBlue{3.37} $&$ 50.03 $&$ 23.95 $&$ 4.28 $&$ 3.10 $&$ 5.67 $&$ \boldBlue{2.64} $\\            
% \hline                
\end{tabular}}                                                                                   
\caption{Test error rates (in \%) with a compression factor of \(\frac{1}{64}\) across all data sets. Best results are printed in \boldBlue{blue}.}
\label{tab:table_2}                                                                         
\end{table*}

%% file: content/discussion.tex
%!TEX root=../hashnn_main.tex
\section{Conclusion}

% \kilian{Point out that 64x compression is equivalent storage to reduced precision with only 1 bit per weight.}

Prior work shows that weights learned in neural networks can be highly redundant~\cite{denil2013predicting}. \abbrev{} exploit this property to create neural networks with ``virtual'' connections that seemingly exceed the storage limits of the trained model. This can have surprising effects. Figure~\ref{fig:fixed_storage} in Section~\ref{sec:exp} shows the test error of neural networks can drop nearly $50\%$, from $3\%$ to $1.61\%$, through expanding the number of weights ``virtually'' by a factor $8\times$. 
Although the collisions (or weight-sharing) might serve as a form of regularization, we can probably safely ignore this effect as both networks (with and without expansion) were also regularized with dropout~\cite{srivastava2014dropout} and the hyper-parameters were carefully fine-tuned through Bayesian optimization. 

So why should additional virtual layers help? One answer is that they probably truly increase the expressiveness of the neural network. As an example, imagine we are provided with a neural network with $100$ hidden nodes. The internal weight matrix has $10000$ weights. If we add another set of $m$ hidden nodes, this increases the expressiveness of the network. If we require all weights of connections to these $m$ additional nodes to be ``re-used'' from the set of existing weights, it is not a strong restriction given the large number of weights in existence. In addition, the backprop algorithm can adjust the shared weights carefully to have useful values for all their occurrences.

As future work we plan to further investigate model compression for neural networks. One particular direction of interest is to optimize \abbrev{} for GPUs. GPUs are very fast (through parallel processing) but usually feature small on-board memory. We plan to investigate how to use \abbrev{} to fit larger networks onto the finite memory of GPUs. A specific challenge in this scenario is to avoid non-coalesced memory accesses due to the pseudo-random hash functions---a sensitive issue for GPU architectures.
%Finally, we plan to incorporate \abbrev{} into the open-source Torch library~\cite{collobert2011torch7}. Particularly interesting are the Torch-iOS\footnote{\url{https://github.com/clementfarabet/torch-ios}} and Torch-Android\footnote{\url{https://github.com/soumith/torch-android}} branches, which will make the benefits of \abbrev{} freely available to application developers.
% around the world.